\documentclass[sn-mathphys-num]{sn-jnl}

\usepackage{natbib} 
\usepackage{float} 
\usepackage{mathptmx}
\usepackage{mathrsfs}
\usepackage{subcaption}

\usepackage{graphicx}%
\usepackage{multirow}%
\usepackage{amsmath}
\usepackage{amssymb}
\usepackage{amsfonts}%
\usepackage{amsthm}%
\usepackage{mathrsfs}%
\usepackage[title]{appendix}%
\usepackage{xcolor}%
\usepackage{textcomp}%
\usepackage{manyfoot}%
\usepackage{booktabs}%
\usepackage{algorithm}%
\usepackage{algorithmicx}%
\usepackage{algpseudocode}%
\usepackage{listings}%
\usepackage{soul}
\usepackage{amsmath, amssymb}

\usepackage{soul}
\usepackage{multirow}
\usepackage[]{mdframed}
\usepackage{xcolor}

\begin{document}

\title[HorNets]{HorNets: Learning from Discrete and Continuous Signals with Routing Neural Networks}


\author*[1,2]{\fnm{Boshko} \sur{Koloski}}\email{boshko.koloski@ijs.si}
\equalcont{These authors contributed equally to this work.}

\author[1]{\fnm{Nada} \sur{Lavra\v{c}}}\email{nada.lavrac@ijs.si}

\author*[1]{\fnm{Bla\v{z}} \sur{\v{S}krlj}}\email{blaz.skrlj@ijs.si}
\equalcont{These authors contributed equally to this work.}

\affil[1]{\orgname{Jo\v{z}ef Stefan Institute}, \orgaddress{\street{Jamova cesta 39}, \city{Ljubljana}, \postcode{1000}, \country{Slovenia}}}
\affil[2]{\orgname{Jo\v{z}ef Stefan Postgraduate School}, \orgaddress{\street{Jamova cesta 39}, \city{Ljubljana}, \postcode{1000}, \country{Slovenia}}}


\abstract{Construction of neural network architectures suitable for learning from both continuous and discrete tabular data is a challenging research endeavor. Contemporary high-dimensional tabular data sets are often characterized by a relatively small instance count, requiring data-efficient learning. We propose HorNets (Horn Networks), a neural network architecture with state-of-the-art performance on synthetic and real-life data sets from scarce-data tabular domains. HorNets are based on a clipped polynomial-like activation function, extended by a custom discrete-continuous routing mechanism that decides which part of the neural network to optimize based on the input's cardinality. By explicitly modeling parts of the feature combination space or combining whole space in a linear attention-like manner, HorNets dynamically decide which mode of operation is the most suitable for a given piece of data with no explicit supervision. This architecture is one of the few approaches that reliably retrieves logical clauses (including noisy XNOR) and achieves state-of-the-art classification performance on 14 real-life biomedical high-dimensional data sets.
HorNets are made freely available under a permissive license alongside a synthetic generator of categorical benchmarks.}


\maketitle
\section{Introduction}
\label{sec:introduction}

In recent years, neural network models have shown great promise when modeling complex real-life data sets such as images, texts, and networks \cite{dong2021survey}. Data sets are most frequently represented in the form of propositional tables. Bader et al. \cite{bader2007} discussed the link between propositional logic and feedforward neural networks. They have successfully demonstrated that non-conventional activations taking values on the interval $[-1, 1]$ need to be adopted to model simple propositional expressions. They also discuss how such expressions can be composed into search trees and binary decision diagrams. Similar ideas were explored by Al Iqbal \cite{al2011eclectic}, where decision trees were co-trained with neural networks to extract simple rules. The more recent work of Labaf et al. \cite{labaf2017propositional} also explored how background knowledge can be incorporated into the learning of logical expressions by approximating the decision process of neural networks with knowledge-aware logical rule learners.

In parallel, this work is also inspired to a large extent by the recent progress in \emph{neuro-symbolic machine learning}. For example,
the works of Amizadeh et al. \cite{amizadeh2020neuro-symbolic} explored the link between logical induction, coupled to low-level (neural) pattern recognition. Similarly, works like DeepProbLog \cite{manhaeve2018deepproblog} explore the language design required to bridge the two learning paradigms (i.e. extension of Prolog with a neural predicate).

Such insights are key motivators of this study, as we show that neural networks can be used to produce, and operate directly with \emph{interpretable} logical expressions. The main contributions of this paper are the following:
\begin{mdframed}
\begin{itemize}
    \item We propose HorNets (Horn Networks), a neural network architecture capable of operating with Horn clauses, that demonstrably achieves state-of-the-art performance on synthetic discrete/categorical and continuous data sets.
    \item HorNets are based on polyClip activation function -- an activation for which we show a probabilistic link to zero-order (propositional) logic.
    \item  To assess the HorNets capability to model logical gates, we conducted extensive benchmarks on synthetic data obtained by a generator implemented as part of this work. We demonstrate that HorNets neural network architecture achieves the best recall in terms of being able to model the most common logic gates (and with that, binary interactions).
    \item Competitive performance of HorNets is demonstrated on more than ten real-life data sets from the biomedical domain, demonstrating it outperforms strong approaches such as gradient-boosted trees and selected AutoML-based classifiers.
\end{itemize}
\end{mdframed}
The rest of this paper is structured as follows. We begin by discussing the related work in Section~\ref{sec:related}, followed by the proposed methodology and its theoretical implications in Section~\ref{sec-hornets-arch}. In Section~\ref{sec:empirical}, we present the considered experimental setting (on synthetic data in Section \ref{sec-empirical-synth} and on real data in Section \ref{sec-eval-real})
, followed by the results in Section~\ref{sec:results}. Finally, in Section \ref{sec:discussion}, we discuss the achievements and drawbacks of this work and present future research directions.

\section{Related work}
\label{sec:related}

This section overviews some of the related work that impacted the development of HorNets. We begin by discussing binarized neural networks, followed by the body of literature focusing on \emph{neuro-symbolic learning}.

Exploration of how rule lists can be transformed into networks capable of classification was performed by \citep{towell1994knowledge}. Their networks could account for AND, OR and NOT connectives, offering expressions in full propositional logic. The subfield of studying neural networks that also resonates with the proposed approach concerns binary neural networks. For example, the work of Rastegari et al. \cite{rastegari2016xnor}  demonstrates that a neural network capable of approximating the XNOR truth table offers sufficient performance on ImageNet, indicating high expressiveness of such networks -- by being able to model XNOR, this was one of the first approaches that demonstrated capability to model general logical statements.
Similar results were achieved on CIFAR-10 and MNIST data sets  \cite{courbariaux2015binaryconnect}.
Binary neural networks were also shown as a promising variant of models that can be computed on specialized hardware such as FPGAs \cite{moss2017high,sun2018xnor}.

Alternatively to considering neural networks that directly approximate logical expressions, such expressions can be generated before being fed into a neural network. This second branch of methods that inspired this work revolves around the notion of \emph{neuro-symbolic} learning. This paradigm explores how neural network-based low-level recognition (e.g., vision) can be coupled with the notion of \emph{reasoning}. 
Such ideas culminated into the concept known as  \emph{deep relational machines} \cite{lodhi2013deep,dash2018large}, offering a state-of-the-art performance for e.g., molecule classification tasks. DeepProbLog \cite{deeprpoblog} combined the idea of probabilistic logic and deep learning, enabling probabilistic inference over deep latent structures, thus providing a tool for tasks such as distant supervision tasks. DeepProbLog \cite{deeprpoblog} integrates probabilistic logic with deep learning, enabling inference over deep latent structures. This framework is ideal for tasks that require background knowledge for validation and utilize deep learning for representation learning in distant supervision scenarios. 
Contrary to the work of \cite{bader2007}, where programs were extracted from feedforward neural networks, an inductive logic program was first used to construct features, which were used to train a deep neural network model. More recently, the link between deep learning and propositionalization was explored by \cite{Lavrac2020}, resulting in two fundamentally different ways to learn latent representations from a propositionalized relational database. \citeauthor{cropper2022inductive} \cite{cropper2022inductive} recently a deeper overview on inductive logic programming and its connections to neuro-symbolic methods.

The recent work on closing the loop between recognition and reasoning \cite{li2020ngs} introduced a grammar model as a symbolic prior to bridge neural perception and symbolic reasoning alongside a top-down, human-like induction procedure. This work demonstrated that such a combined approach significantly outperforms the conventional reinforcement learning-based baselines. The Microsoft research division recently explored the interplay between visual recognition and reasoning \cite{amizadeh2020neuro-symbolic}. They introduced a framework to isolate and evaluate the reasoning aspect of visual question answering separately from its perception, followed by a calibration procedure that explores the relation between reasoning and perception. Further, a neuro-symbolic approach to logical deduction was proposed as \emph{Neural Logic Machines} \cite{nlogm}. This architecture was shown to have inductive logic learning capabilities, which were demonstrated on simple tasks such as sorting. Incorporating semantics-aware logical reasoning to understand better network substructure (communities) was recently explored~\cite{skrlj2020SCD}.

Using neural networks as a tool for learning rules has recently gained importance. \cite{beck2021empirical} explored the application of neural networks to learn rule sets for binary classification using the network structure. \citet{qiao2021learning} proposed DR-Net -- a two-layer neural network architecture for learning logical rule sets. The first layer maps directly interpretable if-then rules, while the output layer forms a disjunction of these rules. Regularisation based on sparsity is used to derive simpler but representative logical rules. On the other hand, \citet{enrl} proposed explainable neural rule learning in a differentiable way. The main idea of the approach is the initial construction of atomic propositions (learned as separate operators via semi-supervised learning) and their subsequent evolution to a binary tree topology to express multiple rules via a neural architecture search. Finally, an ensemble of such trees is created and later used for classification by voting. Similar to the work of \cite{beck2021empirical}, it is limited to binary classification only. \citet{rulenet} recently proposed RuleNet -- an approach that builds on the work of DR-Net, with the additional constraint that the rules must be ordered. In rigorous evaluation, they showcase the network's strong performance on synthetic and real-life classification tasks. However, one significant drawback of the proposed method is the expectation that the input is binary, which is unrealistic for most real-world applications. 

Further, a link between combinatorial optimization solvers and deep learning was proposed \cite{Pogancic2020Differentiation}, offering novel insights into solving hard, e.g., graph-based combinatorial problems. In recent works concerning \emph{program synthesis}, the auto-encoding logical programs \cite{alma9920678983002321} as well as the DreamCoder approach \cite{ellis2020dreamcoder} both demonstrated that symbolic representations can be learned (albeit differently), applicable to solving e.g., physics-related equation discovery (DreamCoder) and learning compact relational representations of a given relational structure (auto-encoding logical programs).

Finally, the proposed work bases some of the ideas from the recent advancements in \emph{understanding} of black-box models. Many deep neural network models are inherently black-box, offering little to no insight into the most representative patterns being learned. As such, the recent trend of model \emph{explainability} in terms of \emph{post-hoc} approximations of model decisions is actively studied. Widely used tools for such approximations are e.g., LIME \cite{ribeiro2016should} and SHAP \cite{NIPS2017_7062}. However, both methods are prone to adversarial attacks, making their ubiquitous use questionable \cite{slack2020fooling}. Alternatively to \emph{post-hoc} explanation of predictions, recent attempts such as TabNet \cite{arik2019tabnet} and propositional Self Attention Networks (SANs) \cite{vskrlj2020feature} indicate that highly relevant features can be extracted via the attention and similar mechanisms directly. \citet{liu2024kan} proposed KAN, a network built on the concept that weights should be placed on the edges, with node activations represented as interactions between functions of the edges, inspired by the Kolmogorov-Arnold representation theorem. This approach provides for interpretable networks, although it could be more scalable.

\section{HorNets Architecture Overview}\label{sec-hornets-arch}
We begin the section with an overview of the HorNets architecture, followed by a theoretical analysis of its expressiveness and computational complexity.

\subsection{General overview and intuition}
\label{sec-general-overview}
\begin{figure}[H]
    \centering
    \includegraphics[width=.95\linewidth]{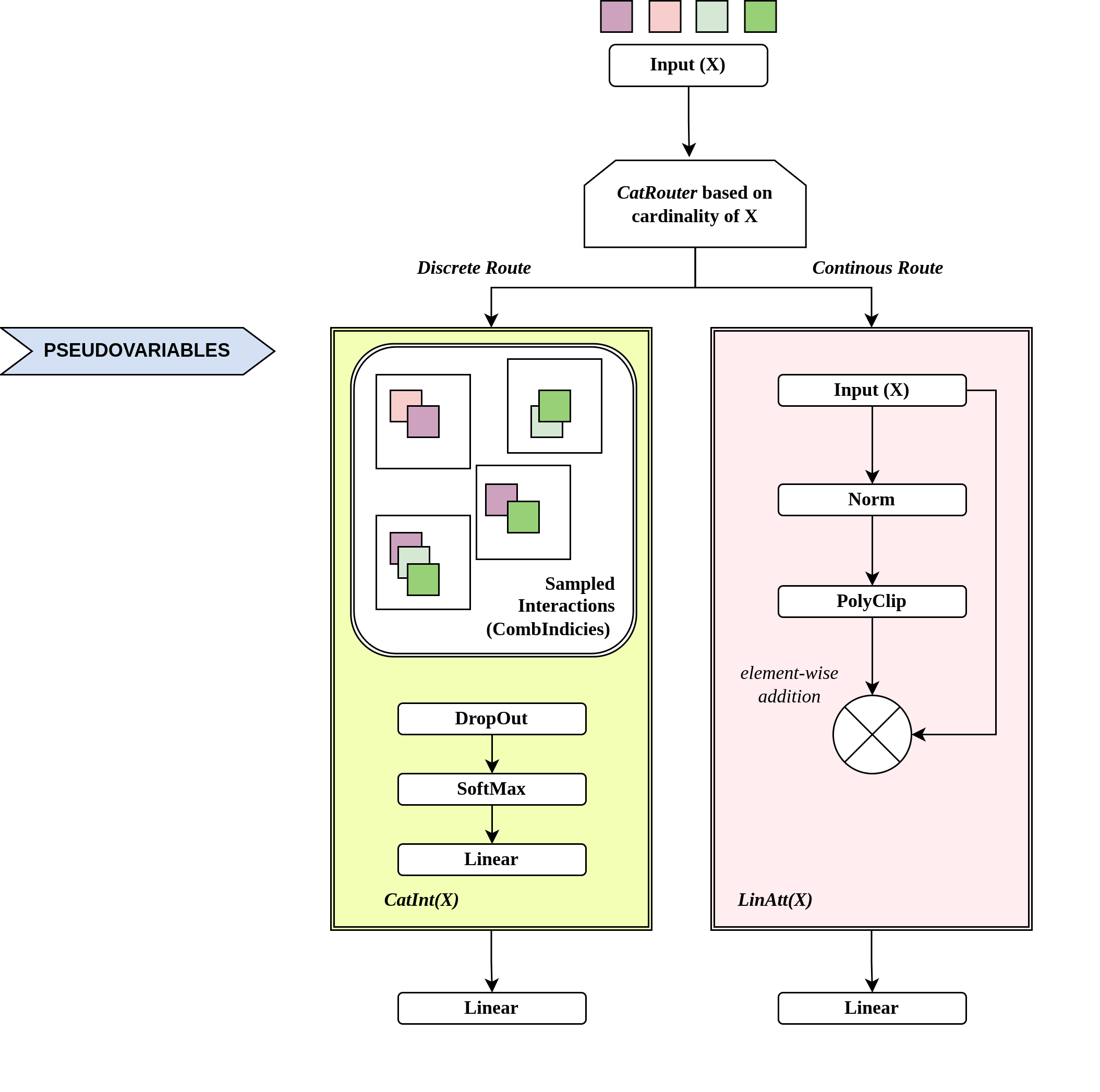}
    \caption{Overview of HorNets architecture. The \emph{CatRouter} (Category Router) component decides whether to treat input as discrete or continuous. If discrete, \emph{CatInt} layer is invoked, conducting factorization-based estimation of a subspace of interactions. If \emph{LinAtt} is invoked, an efficient, element-wise \textsc{PolyClip}-based operation is considered. Both routes end up with a linear layer.}
    \label{fig:overview}
\end{figure}

We begin by discussing a high-level overview of the \textbf{HorNets} architecture, followed by a more detailed inspection in the following sections. The architecture consists of three main components: the \emph{routing operator}, \emph{categorical}, and \emph{continuous blocks}. A schematic overview is shown in Figure~\ref{fig:overview}.
Inputs are batches of tabular data. For each batch, the number of unique values is computed and used to decide which of the subsequent layers to activate. This routing logic is inspired by recent work in the mixture-of-experts-based large language model architectures \cite{cai2024survey}. In this work, all experiments consider categorical layers only if cardinality is lower than 3 (binary matrices). If cardinality exceeds this bound, the continuous part of the architecture takes over. This design choice stems from empirical observation that even if the categorical part of the architecture is substantially scaled, it converges much slower compared to the continuous one -- one of the key purposes of this work is to demonstrate that HorNets is capable of handling (automatically) both scenarios in its current form.

\subsection{Modeling logical operators via activations}
\label{sec:theory}

We first discuss one of the main contributions of this paper, \textsc{polyClip} activation, its formulation and connections to zero-order (propositional) logic.  Conventionally used activations surveyed in \cite{Goodfellow-et-al-2016}, such as ReLU and Sigmoid, offer truncation of weight values to positive real values and were shown to offer state-of-the-art performance on many tasks. On the other hand, activations such as ELU and LReLU offer the inclusion of negative weights; however, they are not symmetric and thus offer little insight into the effect of positive and negative weights. Activations such as \emph{tanH} indeed offer symmetry around zero, and incorporate negative values, however, their infinite bounds hinder interpretability/capability to truncate them to logical clauses directly.

In this work, we introduce the polyClip family of activation functions. For a given parameter $k\in\mathbb N_0$, the corresponding polyClip activation function\footnote{PolyClip -- an acronym for clipped polynomials.} is defined as follows:\\
\begin{mdframed}
\begin{equation}
    \textsc{polyClip}(x, k) = \textrm{clip}(x^{2 \cdot k + 1},-1,1)= \frac{x^{2k+1}}{\max\{|x^{2k+1}|, 1\}} =
    \begin{cases}
    -1 ; x \leq -1\\
    1; x \geq 1\\
    x^{2k+1};\text{else}.
    \end{cases}
\end{equation}
\noindent Here, the $x$ represents real-valued input and $k$ a parameter, determining the order of the considered odd polynomial. 
\end{mdframed}
\vspace{1em}

Note that for the rest of this paper, if $A$ is a real-valued matrix, we denote with $\textrm{polyClip}_k(A)$ the matrix in which the polyClip function is applied to each element.
Apart from being efficient in computing, polyClip activation offers a probabilistic means of expressing statements in formal logic, which we will discuss next.
\begin{figure}[H]
    \centering
    \includegraphics[width = .55\linewidth]{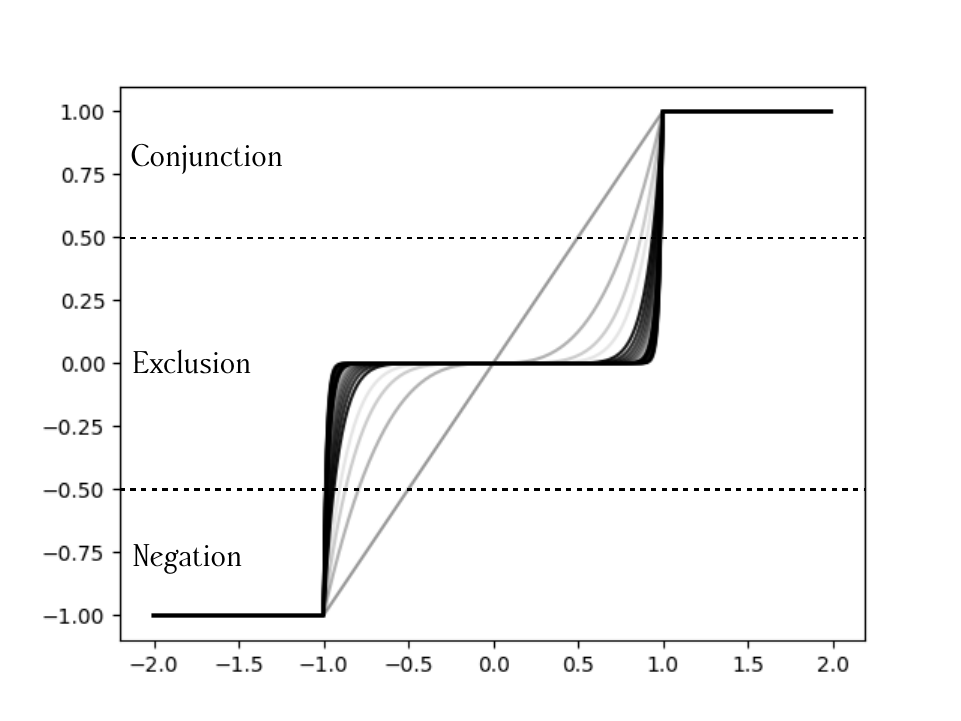}
    \caption{Overview of the \textsc{polyClip} activation and its implications/interpretation when fully discretized (-1, 0, 1) range is considered. The curves were obtained by varying the $k$ parameter.}
    \label{fig:plc}
\end{figure}
The activation projects a weight's value into one of the three main regions, each of which can be associated with the direct interpretation of the \textsc{polyClip}'s effect on the input's value. The three regions of relevance to this work are summarized in Figure~\ref{fig:plc}.
Given an activated weight's value, it can be interpreted as (partial) \emph{negation} (leftmost, negative part of the possible values) and conjunction (rightmost, positive part of the values). Further, values close to 0 imply that a given feature is potentially irrelevant and can be neglected. Multiplication with zero-activated weights in subsequent layers is expected to impact the overall architecture's output less. To maintain the domain and with it the (implicitly) discrete operation on a given input, the main difference between \textsc{polyClip} and the previously mentioned activation functions is \emph{the order} in which it is applied to a given weight and multiplied with a given input\footnote{Note that there exists a scenario where -1, 1 and 0 are outputs of polyClip, enabling direct logical interpretation. In practice, such behavior can be forced with rounding/discretization.}.

We next discuss the theoretical insights, related both to the computational complexity of the main steps in HorNets, as well as its expressiveness and relation to formal logic. We first discuss the behavior of polyClip-activated weights with respect to different types of inputs. 
Let $o$ represent the output of multiplying an activated and \emph{discretized} weight with a given input; thus, $o \in \{-1, 0, 1\}$ -- we assume discretization by \emph{rounding}.
\textsc{polyClip}-activated weights can express negation and conjunction (NOT and AND), and thus model these two types of logical gates. As an example consider a matrix $\boldsymbol{D}$, where each entry is a \emph{discretized} weight from the initial weight matrix, defined as $\boldsymbol{D}_{ij} = round(w_{ij})$. We consider a single hidden-layer neural network-based architecture first. Here, the layer performs an inner product of a column of weights and a row of feature values, i.e. for the $k$-th input vector $\boldsymbol{r}_k$, the output is calculated as
$\sum_{j} \boldsymbol{r}_{kj} \cdot \boldsymbol{D}_{ji}$. In this example, the output will be large (and thus considered during the prediction phase) only if all the summands contribute large values, i.e. if the conjunction of the truth values, to which the individual values $r_{kj} \cdot  \boldsymbol{D}_{ji}$ are mapped, is true, thus showing that polyClip-activated weights can model conjunctions.
Note that the absence (no contribution) of a given feature can also be expressed (\emph{iff} $\boldsymbol{D}_{ij} = 0$).
To show that polyClip-activated weights also express negation, we need to consider the case when $\boldsymbol{D}_{ij} = -1$. In this case, the contributing summand will be large if and only if the $j$-th feature value is negative, i.e. if its 'negation' is true. 
Note that the mentioned relations of activations' values and their impact on output hold true for a single-layer architecture, and can be true (yet it is not guaranteed) for a multi-layer architecture -- interchanging signs can have unpredictable effects in this case. Further, the direct modeling of gates holds true if no other activations are considered simultaneously.
Next, by being able to consider both AND and NOT connectives, we can see that \textsc{polyClip}-activated and discretized weights can model XNOR gate.
Let $\boldsymbol{r}_{kj} \in \{-1,1\}$ and $\boldsymbol{D}_{ji} \in \{-1,1\}$, i.e. each input contributes either positive or negative partial weight value. Next, we establish a mapping where 1 represents True and -1 False. By considering the outputs ($o$) when considering cartesian product of the two input spaces ($\{-1,1\}\times \{-1,1\}$), the following results of the multiplication are possible. The $o = 1$ if both inputs are 1 or -1. If one of the inputs is -1 and the other is 1 (or vice versa), $o = -1$ (without considering the absence of features)

The demonstrated capability illustrates that by considering symmetric positive-negative activation values, clipped to the same range as the input space (-1 and 1 in this case), multiplications between particular values manifest in known gate definitions. Having shown that logical operations can be modeled via discretized activation regime, we can extend the claim to the following.
\textsc{polyClip}-activated weights can express statements in propositional logic.
\begin{mdframed}
Following these definitions, let $\boldsymbol{D}_{ji} = -1$. The inner product $\boldsymbol{r}_{kj} \cdot \boldsymbol{D}_{ji}$ directly translates to $\bigwedge_j \neg (F_j)$. By employing De Morgan's law, $\bigwedge_j \neg F_j \iff \neg (\bigvee_j F_j)$, the conjunctions can be directly expressed. Being able to model conjunction, negation and disjunction, \textsc{polyClip}-activated weights can express statements in disjunctive normal form and hence in propositional logic.
\end{mdframed}
\subsection{HorNets Architecture formulation}
\label{sec:formulation}
We continue the discussion with a more detailed formulation of HorNets architecture. Let $\bf{X} \in \mathbb{R}^{b x d}$ represent an input batch of data ($b$ being batch size and $d$ the dimension). Let $X$ represent the set of values in this batch. The routing layer, named catRouter, denotes a category routing function that distinguishes between two categories: discrete (linAtt -- Linear Attention) and continuous (catInt -- Categorical Interactions). The catRouter mechanism can be in its general form written as
\begin{equation*}
    \textsc{catRouter(X, y)} =
    \begin{cases}
    \textsc{linAtt(X, y)}; |X| > 2;\\
    \textsc{catInt(X, y)}; \textrm{else}.
    \end{cases}
\end{equation*}
The $\textsc{catRouter}(X)$ is thus an efficient operation with negligible practical cost\footnote{Computing cardinality of the types of real-life data sets considered is space-wise negligible -- should exact cardinality computation become a memory bottleneck, probabilistic counting can be employed.}. The subsequent possible operations, $\textsc{linAtt}(x)$ and $\textsc{catInt}(x)$ are discussed next. The $\textsc{linAtt}(x)$ is defined as (recursive formulation aligned with code)
\begin{align*}
 x_0 &= \frac{x}{\max(\lVert x \rVert_p, \epsilon)}\\
 x_1 &= \textsc{polyClip}(x_0) \otimes x \\
 x_2 &= \bf{x_1} \cdot \bf{w} + b.
\end{align*}
This part of the architecture consists of three main steps; normalization $p$ norm ($\epsilon$ is a small constant), activation and elementwise product ($\otimes$), followed by a linear layer. 

The second part of the architecture $\textsc{catRouter}(X)$ can be formulated as follows. Let $\bf{M} \in \mathbb{R}^{o \times |C|}$ represent the \emph{interaction factorization} matrix, where $o$ represents the order of interaction and $C$ the space of input combinations to be considered. This matrix serves as the basis for representing different feature interactions separately, enabling \emph{fine-grained control at interaction level}. The forward pass for this part of the architecture, however, is computationally substantially more expensive and can be formulated as
\begin{align*}
    \bf{F} &= \textsc{combActOp}(\bf{M}, x, \textrm{CombIndices}) \\
    x_0 &= \textsc{dropout(\bf{F}}) \\
    x_1 &= \textsc{softmax}(x_0) \\
    x_2 &= x_1 \cdot \bf{w} + b.
\end{align*}
\begin{mdframed}
The key component is $\textsc{combActOp}$. This operator enables the traversal of considered feature combination space with incremental factorization (and activation). It is defined for a particular feature combination (and its indices) as
\begin{align*}
    \textsc{combActOp}(\bf{M}, x, \textrm{CombIndices}) &= \\
    \bf{F}[:, \textrm{combIndex}] &= \textrm{polyClip}(x[:, \textrm{CombIndices}] \cdot \bf{M}[:, \textrm{CombIndices}]),
\end{align*}
\noindent where CombIndices represent indices mapping to features in the factorization matrix, and CombIndex represents the index of the interaction's representation.\\
\end{mdframed}

The main motivation for explicit factorization of interactions in this component is based on the observation that this type of intermediary factorization substantially simplifies accurate modeling when binary discrete spaces are considered as input. Consider a learning problem of the form $$\{0,1\}^{a \times b} \rightarrow \{0, 1\}^{a}; b > 2.$$ Here $b$ features are used to learn the association between the two spaces. Let us assume target space is defined as $\textsc{XOR}(b_1, b_2)$, i.e. an XOR between the first two features. By considering all interactions of order two, i.e., $\textsc{Comb}(b, 2)$, one of the interactions will also require explicit factorization of representations for $b_1$ and $b_2$. Assuming a neural network is capable of modeling XOR, by being explicitly forced to consider this interaction, it is highly probable it will identify the two features as key to modeling the relation -- linking output to any other feature pair would imply the network learned a wrong interaction that does not generalize. In the context of HorNets, because the considered categorical input space is binary\footnote{Note that one-hot encoding can be used to obtain a binary space from arbitrary categorical inputs.}, interaction factorization will take place as opposed to a linear counterpart that is more suitable for continuous inputs.
In this work, we considered cross-entropy loss for end-to-end optimization of the architecture. The loss is defined as (for a pair of inputs $x$ and $y$).
\begin{equation*}
    \ell(x, y) = - w_{y_n} \log \frac{\exp(x_{n,y_n})}{\sum_{c=1}^C \exp(x_{n,c})}
          \cdot \mathbb{1}\{y_n \not= \text{ignore\_index}\}.
\end{equation*}
Here, $C$ is the set of all classes. Mean aggregation is considered (per batch). 


\subsection{\emph{Pseudovariables} and the curse of higher order interactions}
\label{sec:horder}
While developing the architecture, we observed that unless specific interactions were retrieved, performance on simple tasks such as modeling of logic gates could have been more consistent. By considering higher-order interactions way beyond what is normally considered in matrix factorization (e.g., order 16 and more)\footnote{For high-dimensional real-life data sets, even third-order interactions can be problematic.}, we can force HorNets to sample from larger feature subspaces. However, if e.g., a subspace of 16 interactions contains the two features that govern the output signal, yet the remaining 14 represent pure noise, optimization (backpropagation) was observed to have issues with convergence. To remedy this shortcoming, we introduced the notion of \emph{pseudovariables} -- artificially added features that, in the context of HorNets, serve as mask variables. These variables are initialized as constant space of ones, and don't change the results of intermediary computation. Each HorNets evaluation automatically introduced $k$ pseudovariables, where $k$ is the order of interactions being modeled. The rationale for this design choice is that by being able to mask most of a higher-order interaction, lower-order interactions can be reliably retrieved, while at the same time, fewer samples of the feature space are required, thus reducing overall computational complexity -- this observation is also one of the reasons HorNets are approachable on commodity hardware in its current form\footnote{Note that sampling higher order interactions exhaustively is computationally intractable, even for moderately sized data sets (e.g., hundred features).}. An interesting side-effect of pseudovariables is increased interpretability. As the categorical part of the architecture performs softmax per-interaction, highest-scored interactions can be directly inspected, and serve as possible candidates for underlying rules. \\
\begin{mdframed}
If, for example, a rule of form $f_1 \wedge f_2 \wedge p_3 \wedge p_4$ is obtained, where $p_x$ are pseudovariables, we can reliably claim that only $f_1$ and $f_2$ have the impact on learning (as by definition $p_3$ and $p_4$ contribute nothing).
\end{mdframed}

\textbf{A note on computational complexity}.
Having discussed the architecture, it is worth noting that the two main components between which the router layer decides are subject to notably different computational complexity. In particular the \textsc{linAtt(x, y)} operation is linear with regards to the number of input dimensions, while \textsc{CatInt(x, y)} can be subject to a combinatorial explosion (many features, higher-order interactions), and is thus \emph{exponential in the limit}\footnote{$\mathcal{O}(\textrm{Comb}(|I|, \textrm{order}))$, this can be shown via Sterling's approximation -- the proof can be found in the Appendix}. In practice, however, we observed that a relatively small number of combinations of higher order is sufficient for modeling categorical data well. Many reasons for this are based on empirical evaluation \emph{pseudovariables}. Further, we implemented a Monte Carlo-based sampling of interactions within the network to mitigate for the explosion, effectively changing which interactions are considered per epoch. Hence, sufficiently long training cycles in the limit sample all possible combinations, even though in practice HorNets converge faster. The architecture was built to be practical -- all experiments discussed in the following sections can be replicated on a moderate off-the-shelf laptop with no GPU.

\section{Evaluation}
\label{sec:empirical}
In this section, we set up the evaluation scenario. We focus on assessing the predictive power of our method -- first on synthetic data, where we want to analyze the logic modeling capabilities and second on real-life downstream classification tasks.

\subsection{Empirical evaluation on synthetic data}
\label{sec-empirical-synth}
The first round of experiments revolves around benchmarking of HorNets' logic modeling capabilities. To systematically evaluate this trait, we designed a synthetic data generator that enables the construction of arbitrary-dimensional binary training datasets, where target space is defined as a logical combination of a given feature subspace. We considered the main logic operators, namely AND, OR, NOT, XOR and XNOR. We refer to these gates as $\textsc{OP}$.
The dimension parameter $d$ parametrizes the generator, and the instance count parameter $c$. It first generates a binary matrix $\bf{B} \in \{0, 1\}^{c \times d}$, where $\forall k \in \bf{B} | k \sim \textsc{Bernoulli}(0.5)$. Note that $\textsc{OP}$ are binary operators -- they comprise second-order interactions. For each operator, generator computes the target vector $\bf{t} \in \{0, 1\}^c$ as $t_i = \textsc{OP}(k_{i, j_0}, k_{i, j_1})$. The remainder of the $\bf{B}$ remains random (noise). The higher the $d$, the harder it is estimated that a given classifier can retrieve the explicitly encoded interaction (two features). Note that XNOR and XOR are of particular interest because both inputs must be considered simultaneously (non-myopic model). We generated 30 data sets for each gate and $d$ of interest. Further, $c$ was set to 128 in all synthetic experiments.
To evaluate the relative performance of our method, we compare it with Logistic Regression (LR), Random Forest (RF) \cite{le2020scaling}, TabNet \cite{arik2019tabnet} and an MLP classifier. We consider two variants of our method: one with the ReLU activation function and another with the proposed PolyClip variant.

\subsection{Empirical evaluation on real-life data}
\label{sec-eval-real}
We proceed by an overview of the HorNets performance on real-life biomedical data. 
The motivation for this is twofold: First, biomedical data features an abundant number of variables but a scarce number of examples, which leads to problematic behavior when MLPs are applied directly. Second, neural network-based approaches such as TabNET and MLPs are hardly interpretable, posing challenges for the real-life application of machine learning models in medicine.
The data sets \cite{li2018feature} considered represent different classification problems that aim to associate gene expression signals with the target output (e.g. tumor presence). A summary of data sets is shown in Table \ref{tab:data}.

\begin{table}[t!]
\centering
\caption{Dataset summary. For each dataset, we report the number of instances, features, classes, and class distribution (Class 1, Class 2, and combined distribution of remaining classes).}
\begin{tabular}{lcccccc}
\hline
Dataset & Instances & Features & Classes & C1 & C2  & C-remainder \\
\hline
Multi-B \cite{tissues} & 32 & 5565 & 4 & 34.38 & 28.12 & 37.50 \\
Breast-B \cite{tissues} & 49 & 1213 & 4 & 38.78 & 24.49 & 36.73 \\
DLBCL-C\cite{tissues}& 58 & 3795 & 4 & 29.31 & 27.59 & 43.10 \\
Breast-A \cite{tissues} & 98 & 1213 & 3 & 52.04 & 36.73 & 11.22 \\
Prostate-GE \cite{singh2002gene}& 102 & 5966 & 2 & 50.98 & 49.02 & - \\
Multi-A  \cite{tissues}  & 103 & 5565 & 4 & 27.18 & 25.24 & 47.58 \\
CLL  \cite{haslinger2004microarray} & 111 & 11340 & 3 & 45.95 & 44.14 & 9.91 \\
DLBCL-D \cite{tissues} & 129 & 3795 & 4 & 37.98 & 28.68 & 33.33 \\
DLBCL-A \cite{tissues} & 141 & 661 & 3 & 35.46 & 34.75 & 29.79 \\
TOX \cite{bhattacharjee2001classification} & 171 & 5748 & 4 & 26.32 & 26.32 & 47.36 \\
DLBCL-B \cite{tissues} & 180 & 661 & 3 & 48.33 & 28.33 & 23.33 \\
SMK \cite{spira2007airway} & 187 & 19993 & 2 & 51.87 & 48.13 & - \\
Lung  \cite{bajwa2016cutting} & 203 & 3312 & 5 & 68.47 & 10.34 & 21.18 \\
TCGA  \cite{weinstein2013cancer}  & 801 & 20531 & 5 & 37.45 & 18.23 & 44.32 \\ \hline
\end{tabular}
\label{tab:data}
\end{table}

Motivated by the work of \citep{grinsztajn2022tree}, we aim to compare our method against various classifiers, including linear, ensemble, neural networks, and AutoML classifiers. Specifically, we have selected the following classifiers:
\begin{itemize}
\item 
\texttt{Linear}: Decision Tree (DT), Logistic Regression (Lasso (L1) and Ridge (L2)), and Support Vector Machine (SVM) \cite{cortes1995support}; 
\item \texttt{Ensemble}: Random Forest (RF) \cite{breiman2017classification} and XGBoost (XGB) \cite{chen2016xgboost};
\item \texttt{Neural}: Multi-Layer Perceptron (MLP), TabNET \cite{arik2019tabnet}, and LassoNet \cite{lemhadri2021lassonet}; 
\item \texttt{AutoML}: TPOT \cite{le2020scaling} -- an Automated Machine Learning tool that optimizes machine learning pipelines using genetic programming.
\end{itemize}

All models, except the MLP model, were utilized from their official libraries (scikit-learn for linear and ensemble classifiers, PyTorch-TabNet for TabNET, and TPOT for TPOT). The MLP classifier was implemented using the PyTorch suite, consisting of an input layer followed by a dense layer of 32 neurons, preceding the final classification layer, optimized with the Adam optimizer for up to 100 epochs with early stopping after 10 epochs.
We use the macro F1-score to compare classifiers, conducting 5-fold stratified cross-validation across five different seeds, predicting 25 times for a given dataset to assess the variability of results. We search through the following hyperparameters, reporting the best-performing per dataset:
\texttt{Activation}: ReLU and our proposed PolyClip; \texttt{Order}: 4, 8, 16, 32, 64, 128, 512, 1024, 2048, 4096; \texttt{Number of Rules}: 4, 8, 16, 32, 64, 128, 512, 1024, 2048, 4096; \texttt{Learning Rate}: 0.001, 0.01, 0.1; \texttt{Batch Size}: fixed at 15. 

\section{Results}
\label{sec:results}

This section presents the experimental results of the synthetic and the real biomedical experiments, followed by the analysis of the execution time and impact of hyperparameters on the method.

\subsection{Synthetic results}

Table \ref{tab:res_synth} presents the performance of our method on synthetic logic modeling datasets. Using the PolyClip variant, our method achieves a perfect score, demonstrating its ability to model arbitrary Horn clauses. Strong competitors such as Logistic Regression, Random Forest, and MLP could model logical operations like AND, NOT, OR, and XNOR when the number of variables was less than 8. All models, with the exclusion of the Random Forest, failed to model the XOR operator. Interestingly, the attention-based TabNET method failed to successfully model any interactions regardless of the number of variables across all problems. High-dimensional settings posed challenges for most methods, except for the proposed HorNet. These results highlight the ability of HorNets with the PolyClip variant to model arbitrary logical clauses effectively.

\begin{table}[!t]
        \caption{Overview of F1 performance of different algorithms on a collection of synthetic data sets (30 repetitions, mean and deviation reported).}
    \centering  
    \begin{tabular}{lllllll}
\toprule
Problem & TabNet & Random & MLP & Logistic & HorNets & \textbf{HorNets} \\
 & & Forest & & Regression & -ReLU & \textbf{-PolyClip} \\
\midrule

and(dim=3) & 0.296 $\pm$ 0.06 & 1.0 $\pm$ 0.0 & 1.0 $\pm$ 0.0 & 1.0 $\pm$ 0.0 & 1.0 $\pm$ 0.0 & 1.0 $\pm$ 0.0 \\
and(dim=4) & 0.204 $\pm$ 0.136 & 1.0 $\pm$ 0.0 & 0.905 $\pm$ 0.134 & 1.0 $\pm$ 0.0 & 1.0 $\pm$ 0.0 & 1.0 $\pm$ 0.0 \\
and(dim=16) & 0.512 $\pm$ 0.137 & 0.981 $\pm$ 0.06 & 0.973 $\pm$ 0.063 & 0.981 $\pm$ 0.06 & 0.935 $\pm$ 0.178 & 1.0 $\pm$ 0.0 \\
and(dim=32) & 0.331 $\pm$ 0.147 & 0.901 $\pm$ 0.122 & 0.899 $\pm$ 0.131 & 0.912 $\pm$ 0.107 & 0.821 $\pm$ 0.21 & 1.0 $\pm$ 0.0 \\
and(dim=8) & 0.465 $\pm$ 0.089 & 0.978 $\pm$ 0.069 & 0.981 $\pm$ 0.06 & 1.0 $\pm$ 0.0 & 0.941 $\pm$ 0.187 & 1.0 $\pm$ 0.0 \\ 
and(dim=64) & 0.584 $\pm$ 0.151 & 0.783 $\pm$ 0.205 & 0.779 $\pm$ 0.21 & 0.805 $\pm$ 0.191 & 0.95 $\pm$ 0.083 & 1.0 $\pm$ 0.0 \\ 
and(dim=128) & 0.463 $\pm$ 0.085 & 0.484 $\pm$ 0.093 & 0.655 $\pm$ 0.178 & 0.627 $\pm$ 0.185 & 0.811 $\pm$ 0.209 & 1.0 $\pm$ 0.0 \\
\hline

not(dim=3) & 0.446 $\pm$ 0.023 & 1.0 $\pm$ 0.0 & 1.0 $\pm$ 0.0 & 1.0 $\pm$ 0.0 & 1.0 $\pm$ 0.0 & 1.0 $\pm$ 0.0 \\
not(dim=4) & 0.0 $\pm$ 0.0 & 1.0 $\pm$ 0.0 & 1.0 $\pm$ 0.0 & 1.0 $\pm$ 0.0 & 1.0 $\pm$ 0.0 & 1.0 $\pm$ 0.0 \\
not(dim=8) & 0.629 $\pm$ 0.256 & 0.948 $\pm$ 0.164 & 0.948 $\pm$ 0.164 & 1.0 $\pm$ 0.0 & 1.0 $\pm$ 0.0 & 1.0 $\pm$ 0.0 \\ 
not(dim=16) & 0.418 $\pm$ 0.036 & 1.0 $\pm$ 0.0 & 0.896 $\pm$ 0.219 & 1.0 $\pm$ 0.0 & 1.0 $\pm$ 0.0 & 1.0 $\pm$ 0.0 \\
not(dim=32) & 0.146 $\pm$ 0.094 & 1.0 $\pm$ 0.0 & 1.0 $\pm$ 0.0 & 1.0 $\pm$ 0.0 & 1.0 $\pm$ 0.0 & 1.0 $\pm$ 0.0 \\
not(dim=64) & 0.435 $\pm$ 0.041 & 1.0 $\pm$ 0.0 & 1.0 $\pm$ 0.0 & 1.0 $\pm$ 0.0 & 1.0 $\pm$ 0.0 & 1.0 $\pm$ 0.0 \\
not(dim=128) & 0.521 $\pm$ 0.169 & 1.0 $\pm$ 0.0 & 1.0 $\pm$ 0.0 & 1.0 $\pm$ 0.0 & 1.0 $\pm$ 0.0 & 1.0 $\pm$ 0.0 \\
\hline
or(dim=3) & 0.255 $\pm$ 0.135 & 1.0 $\pm$ 0.0 & 1.0 $\pm$ 0.0 & 1.0 $\pm$ 0.0 & 1.0 $\pm$ 0.0 & 1.0 $\pm$ 0.0 \\
or(dim=4) & 0.574 $\pm$ 0.211 & 1.0 $\pm$ 0.0 & 0.961 $\pm$ 0.125 & 1.0 $\pm$ 0.0 & 1.0 $\pm$ 0.0 & 1.0 $\pm$ 0.0 \\
or(dim=8) & 0.265 $\pm$ 0.116 & 1.0 $\pm$ 0.0 & 1.0 $\pm$ 0.0 & 1.0 $\pm$ 0.0 & 1.0 $\pm$ 0.0 & 1.0 $\pm$ 0.0 \\ 
or(dim=16) & 0.399 $\pm$ 0.101 & 0.988 $\pm$ 0.039 & 0.99 $\pm$ 0.031 & 0.988 $\pm$ 0.039 & 1.0 $\pm$ 0.0 & 1.0 $\pm$ 0.0 \\
or(dim=32) & 0.48 $\pm$ 0.175 & 0.91 $\pm$ 0.123 & 0.87 $\pm$ 0.168 & 0.92 $\pm$ 0.078 & 1.0 $\pm$ 0.0 & 1.0 $\pm$ 0.0 \\
or(dim=64) & 0.34 $\pm$ 0.081 & 0.741 $\pm$ 0.201 & 0.73 $\pm$ 0.197 & 0.748 $\pm$ 0.215 & 1.0 $\pm$ 0.0 & 1.0 $\pm$ 0.0 \\
or(dim=128) & 0.285 $\pm$ 0.073 & 0.558 $\pm$ 0.244 & 0.764 $\pm$ 0.199 & 0.699 $\pm$ 0.227 & 1.0 $\pm$ 0.0 & 1.0 $\pm$ 0.0 \\

\hline

xnor(dim=3) & 0.547 $\pm$ 0.24 & 1.0 $\pm$ 0.0 & 1.0 $\pm$ 0.0 & 1.0 $\pm$ 0.0 & 1.0 $\pm$ 0.0 & 1.0 $\pm$ 0.0 \\
xnor(dim=4) & 0.093 $\pm$ 0.067 & 1.0 $\pm$ 0.0 & 1.0 $\pm$ 0.0 & 1.0 $\pm$ 0.0 & 1.0 $\pm$ 0.0 & 1.0 $\pm$ 0.0 \\
xnor(dim=8) & 0.729 $\pm$ 0.286 & 1.0 $\pm$ 0.0 & 0.948 $\pm$ 0.164 & 1.0 $\pm$ 0.0 & 1.0 $\pm$ 0.0 & 1.0 $\pm$ 0.0 \\ 
xnor(dim=16) & 0.4 $\pm$ 0.047 & 1.0 $\pm$ 0.0 & 0.948 $\pm$ 0.164 & 1.0 $\pm$ 0.0 & 1.0 $\pm$ 0.0 & 1.0 $\pm$ 0.0 \\
xnor(dim=32) & 0.19 $\pm$ 0.086 & 1.0 $\pm$ 0.0 & 0.948 $\pm$ 0.164 & 1.0 $\pm$ 0.0 & 1.0 $\pm$ 0.0 & 1.0 $\pm$ 0.0 \\
xnor(dim=64) & 0.543 $\pm$ 0.244 & 1.0 $\pm$ 0.0 & 1.0 $\pm$ 0.0 & 1.0 $\pm$ 0.0 & 1.0 $\pm$ 0.0 & 1.0 $\pm$ 0.0 \\
xnor(dim=128) & 0.623 $\pm$ 0.261 & 1.0 $\pm$ 0.0 & 1.0 $\pm$ 0.0 & 1.0 $\pm$ 0.0 & 1.0 $\pm$ 0.0 & 1.0 $\pm$ 0.0 \\
\hline

xor(dim=3) & 0.483 $\pm$ 0.126 & 1.0 $\pm$ 0.0 & 0.688 $\pm$ 0.141 & 0.535 $\pm$ 0.211 & 1.0 $\pm$ 0.0 & 1.0 $\pm$ 0.0 \\
xor(dim=4) & 0.511 $\pm$ 0.18 & 1.0 $\pm$ 0.0 & 0.747 $\pm$ 0.15 & 0.497 $\pm$ 0.189 & 1.0 $\pm$ 0.0 & 1.0 $\pm$ 0.0 \\
xor(dim=8) & 0.326 $\pm$ 0.12 & 0.975 $\pm$ 0.041 & 0.992 $\pm$ 0.025 & 0.459 $\pm$ 0.106 & 0.984 $\pm$ 0.033 & 1.0 $\pm$ 0.0 \\
xor(dim=16) & 0.42 $\pm$ 0.104 & 0.842 $\pm$ 0.099 & 0.913 $\pm$ 0.077 & 0.465 $\pm$ 0.169 & 0.975 $\pm$ 0.081 & 1.0 $\pm$ 0.0 \\
xor(dim=32) & 0.436 $\pm$ 0.125 & 0.678 $\pm$ 0.158 & 0.647 $\pm$ 0.237 & 0.554 $\pm$ 0.188 & 0.911 $\pm$ 0.172 & 1.0 $\pm$ 0.0 \\
xor(dim=64) & 0.471 $\pm$ 0.09 & 0.451 $\pm$ 0.081 & 0.522 $\pm$ 0.162 & 0.476 $\pm$ 0.146 & 0.908 $\pm$ 0.184 & 1.0 $\pm$ 0.0 \\
xor(dim=128) & 0.45 $\pm$ 0.097 & 0.473 $\pm$ 0.146 & 0.505 $\pm$ 0.143 & 0.584 $\pm$ 0.107 & 0.966 $\pm$ 0.082 & 1.0 $\pm$ 0.0 \\
\bottomrule
\end{tabular}
      \label{tab:res_synth}
\end{table}

\pagebreak
\subsection{Real data sets}

\begin{figure}[H]
    \centering
    \includegraphics[width=0.9\linewidth]{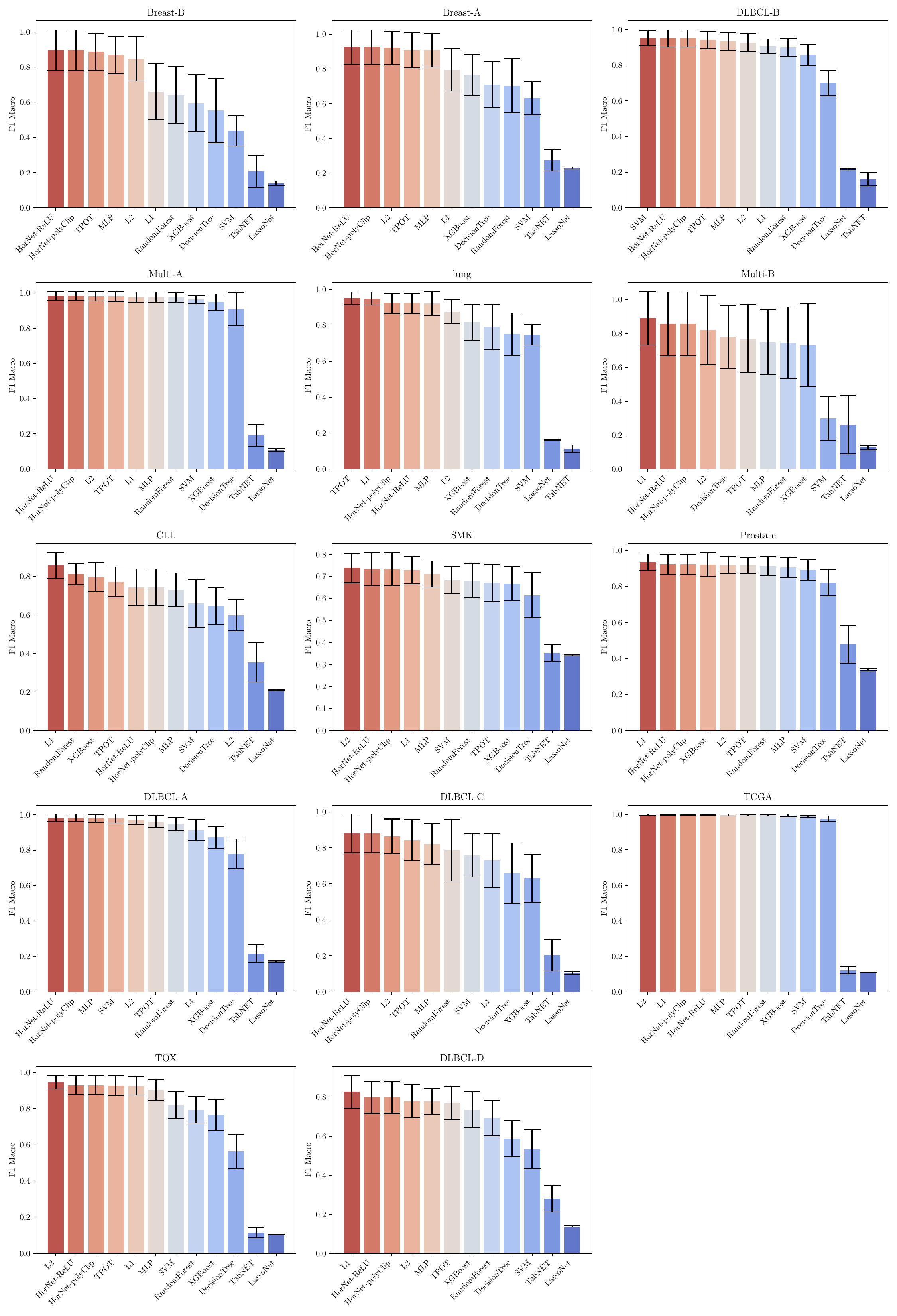}
    \caption{Overview of algorithm performance (14 data sets). It can be observed that HorNets consistently rank among the top three performers, indicating the capability of this algorithm to operate on different problems consistently. Its main competitor in terms of tabular neural network learning, TabNet, failed to achieve high performance -- it is hypothesized this is due to high dimensionality/multiple classes of data considered.}
    \label{fig:algo-perf}
\end{figure}

We show the results on the real biomedical experiments runs in Figure \ref{fig:algo-perf}.

We notice that both variants of the HorNet -- the one based on the polyClip activation and the one based on the ReLU activation surpass the performance on 5 out of 14 datasets, surpassing the performance of the state-of-the art models like the TPOT and the L1 and L2 models.

\begin{figure}[H]
    \centering
    \includegraphics[width=\linewidth]{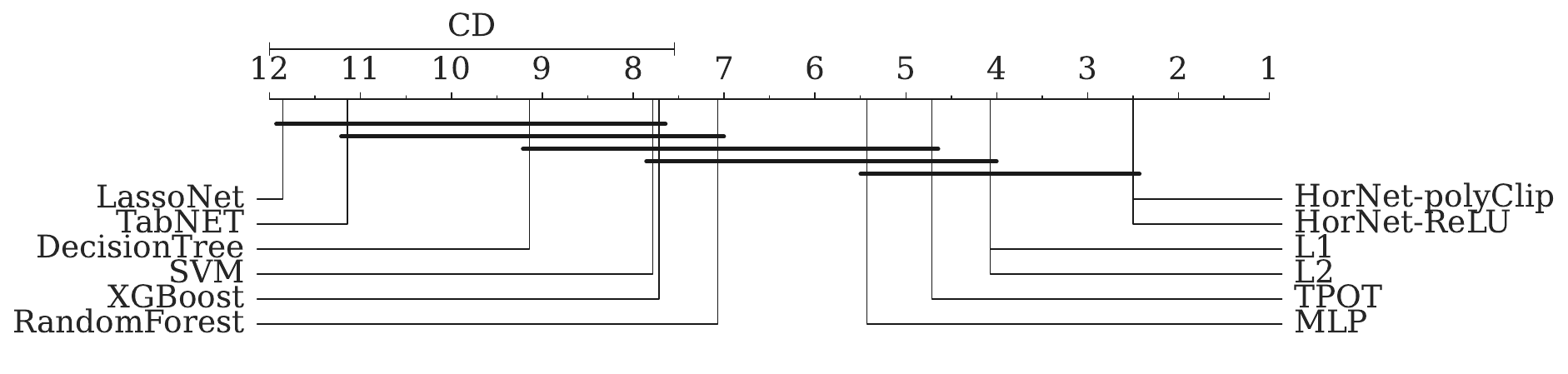}
    \caption{Friedman-Nemenyi test demonstrating the competitive performance of our method's F1-macro metric.}
    \label{fig:algo-test}
\end{figure}

We further investigate the statistical difference between these models by conducting Friedman-Nemenyi test \cite{demvsar2006statistical} in Figure \ref{fig:algo-test}. The results show that both variants perform similarly, outperforming the remaining models and performing within critical distance to the best-ranked ones -- L1, L2, TPOT and MLP models.  

\begin{figure}[H]
    \centering
    \resizebox{0.5\textwidth}{!}{\includegraphics{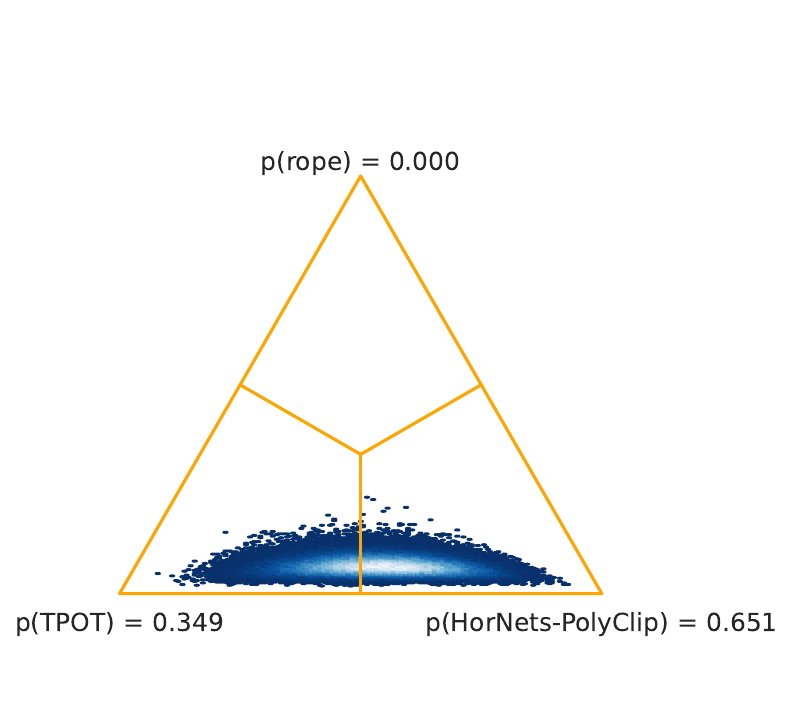}}
    \caption{Bayesian hierarchical t-test assessing differences between the HorNets-PolyClip variant and the TPOT AutoML model. The test indicates insignificant differences between HorNets-PolyClip and the state-of-the-art AutoML approach. The proposed approach is even marginally better.}
    \label{fig:bayesiantest}
\end{figure}

The Bayesian Hierarchy test \cite{JMLR:v18:16-305} between our model and the TPOT model shown in Figure \ref{fig:bayesiantest} also shows that the model is significantly better with a probability of 65\%, while the TPOT model is better with a probability of 35\% (within a region-of-practical-equivalence of 0.01).

\subsection{Execution time}
\begin{figure}[H]
    \centering
    \resizebox{\textwidth}{!}{\includegraphics{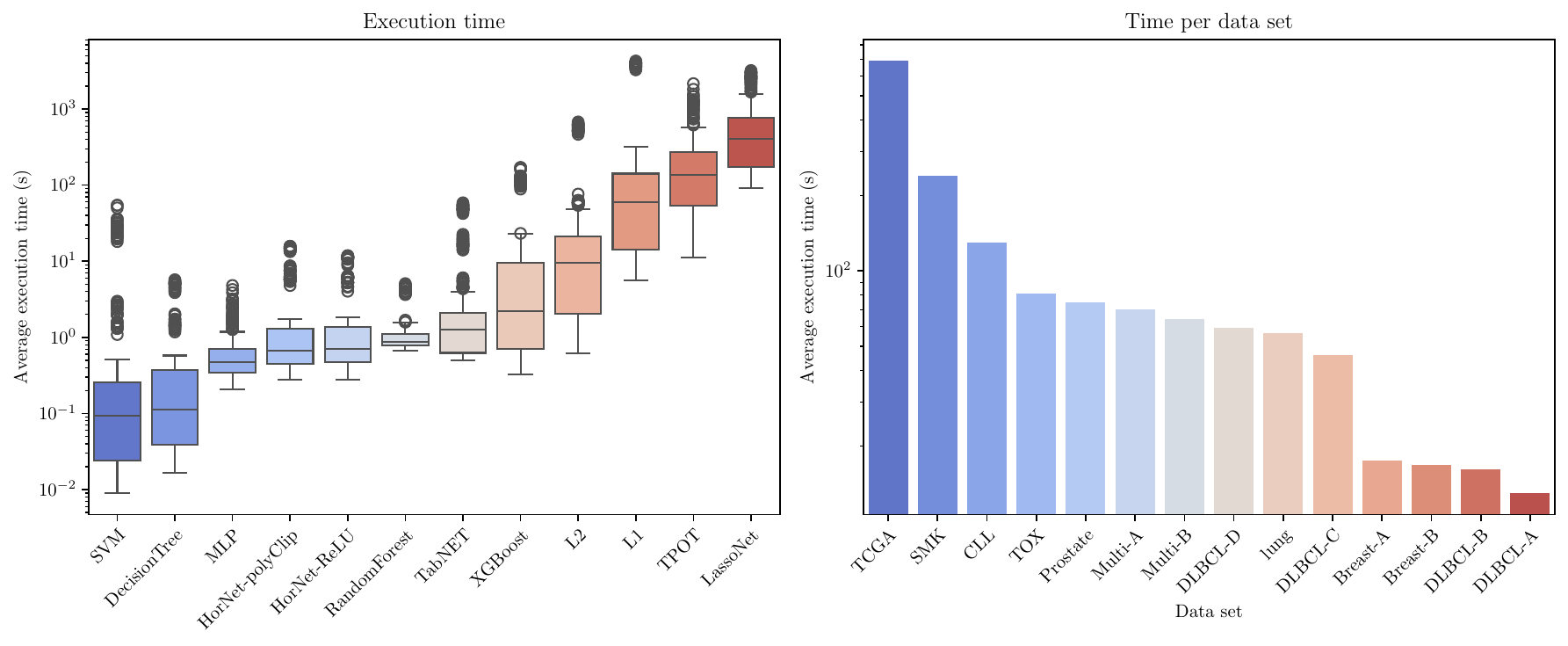}}
    \caption{Execution time comparison of our method against the baselines.}
    \label{fig:exec-time}
\end{figure}

Next, we analyze the execution time of the methods shown in Figure \ref{fig:exec-time}. Our results show that both the polyClip and ReLU variants exhibit indistinguishable performance. The proposed method significantly outperforms advanced neural baselines such as LassoNet and TPOT (which exhaustively search for solutions). It also outperforms ensemble methods such as RandomFrost and XGBoost. Furthermore, we find that the execution time of all methods increases with the size of the dataset, both in terms of the number of features and the number of samples.

\begin{figure}[H]
    \centering
    \resizebox{\textwidth}{!}{\includegraphics{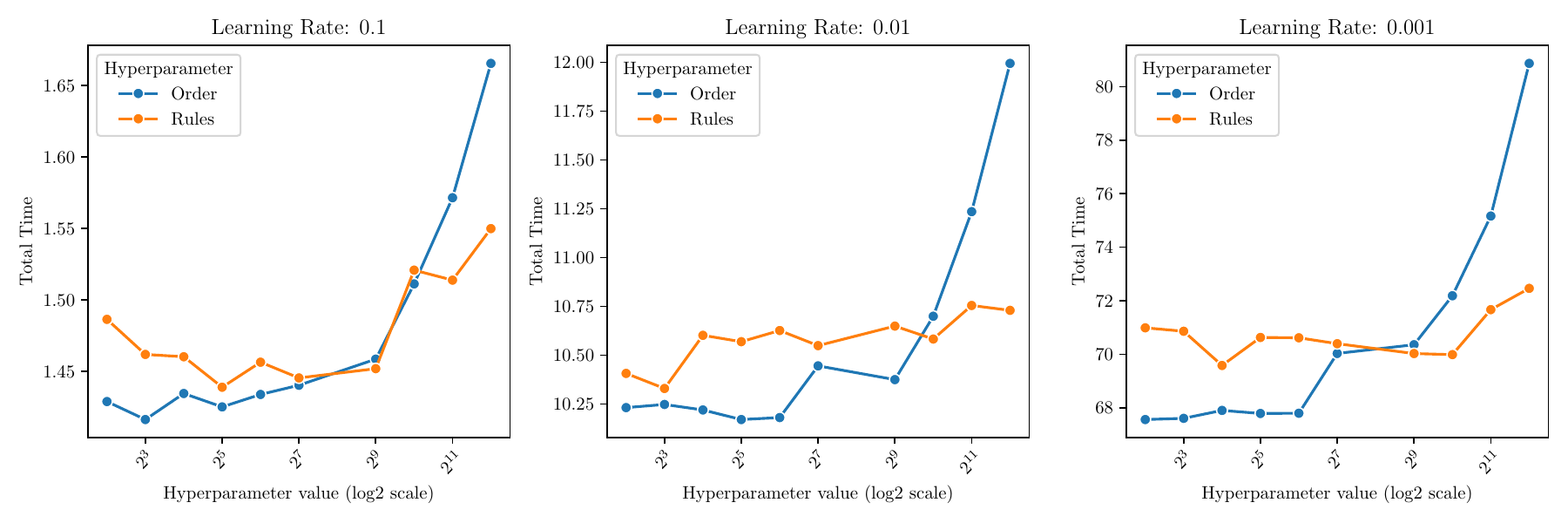}}

    \caption{Effect of different hyper-parameters on execution time (seconds) of our method.}
    \label{fig:exec-time-params}
\end{figure}

In Figure \ref{fig:exec-time-params} we analyze the effects of the learning rates and the different number of parameters on the execution time. The empirical results confirm the theoretical bounds on the execution time, with the order parameter increasing exponentially and the number of rules increasing linearly. 

\subsection{Impact of different hyperparameters}
\begin{figure}[H]
    \centering
    \resizebox{\textwidth}{!}{0.47\includegraphics{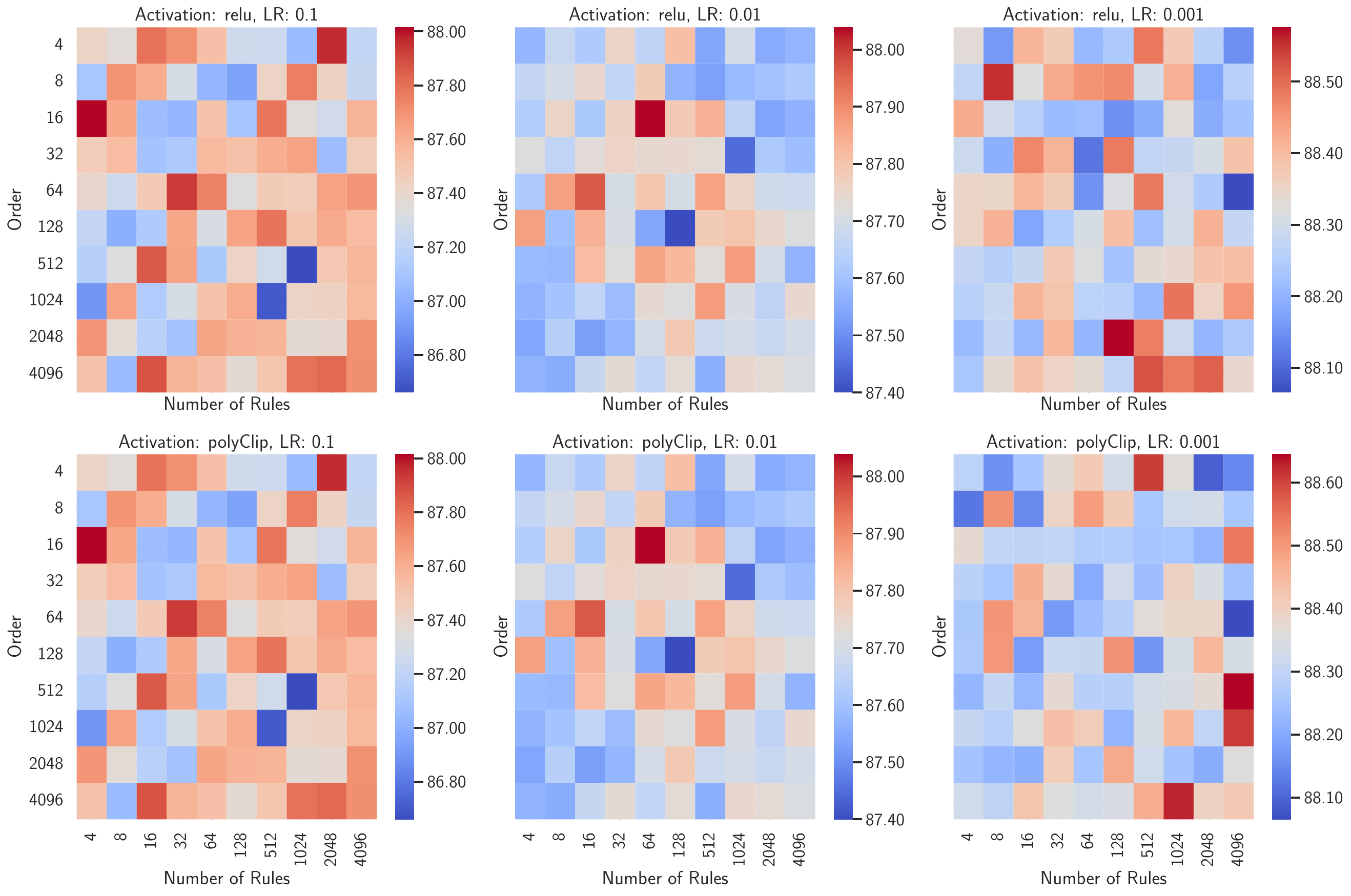}}
    \caption{Interaction of different order and number of rules at different learning rates. Results indicate learning rate is the key parameter that has to be tuned -- if an appropriate value is selected, relatively low order/rule count are sufficient for good performance.}
    \label{fig:heatmap-interact}
\end{figure}

Next, we analyze the impact of different hyperparameters. Recall, we use the following grid to search and evaluate for hyperparameters \texttt{Activation}: ReLU and our proposed PolyClip; \texttt{Order}: 4, 8, 16, 32, 64, 128, 512, 1024, 2048, 4096; \texttt{Number of Rules}: 4, 8, 16, 32, 64, 128, 512, 1024, 2048, 4096; \texttt{Learning Rate}: 0.001, 0.01, 0.1; \texttt{Batch Size}: fixed at 15.  For each tuple of parameters, we train on all datasets in the described 5 run 5-fold cross-validation setting. Figure \ref{fig:heatmap-interact} shows the optimization landscape across different hyperparameter settings, showing the interaction of different activation functions, the learning rates with the order and number of rules. The results show that no clear best setting can be found and that a theoretical parameter sweep is required when trying to optimize for the best results. Note that, however, the difference between the best-performing (88.60\%) and worst-performing parameter setting (86.10\%) on average is 1.98 percentage points, which we hypothesize.  

However, we find that there is no significant variance in the results between the different datasets. In Table \ref{table:top5} we present the top-5 performing parameter settings. The results indicate a negligible difference in performance across these top-5 configurations, despite variations in the combination of order and rules. As expected, a smaller learning rate consistently yielded the highest F1-scores. Table \ref{tab:worst5} showcases that a similar pattern is observed in the analysis of the top-5 worst-performing parameter settings. Here too, the differences between these combinations are minimal. Notably, the higher learning rates were associated with poorer results, reinforcing the notion that a lower learning rate is preferable for achieving better performance. Notably, the variability within datasets is consistently about 12.5 percentage points regardless of the parameter, pointing at the stability of the method.

\begin{table}[!htb]
    \centering
        \caption{Top 5 Best-Performing Parameters.}
        \label{table:top5}
        \begin{tabular}{lrrrl}
        \toprule
        Activation & Learning Rate & Order & Rules & F1-score \\
        \midrule
        polyClip & 0.001 & 512 & 4096 & 88.65 ± 12.40 \\
        polyClip & 0.001 & 4096 & 1024 & 88.63 ± 12.31 \\
        polyClip & 0.001 & 1024 & 4096 & 88.61 ± 12.36 \\
        polyClip & 0.001 & 4 & 512 & 88.61 ± 12.40 \\
        ReLU & 0.001 & 2048 & 128 & 88.58 ± 12.44 \\
        \bottomrule
        \end{tabular}
    \label{tab:top5}
\end{table}

\begin{table}[!htb]
    \centering
        \caption{Bottom 5 Worst-Performing Parameters.}
        \begin{tabular}{lrrrl}
        \toprule
        Activation & Learning Rate & Order & Rules & F1-score \\
        \midrule
        polyClip & 0.10 & 1024 & 4 & 86.90 ± 12.72 \\
        ReLU & 0.10 & 1024 & 512 & 86.70 ± 13.07 \\
        polyClip & 0.10 & 1024 & 512 & 86.70 ± 13.07 \\
        polyClip & 0.10 & 512 & 1024 & 86.66 ± 13.19 \\
        ReLU & 0.10 & 512 & 1024 & 86.66 ± 13.19 \\
        \bottomrule
        \end{tabular}
    \label{tab:worst5}
\end{table}

\begin{figure}[H]
    \centering
    \resizebox{\textwidth}{!}{\includegraphics{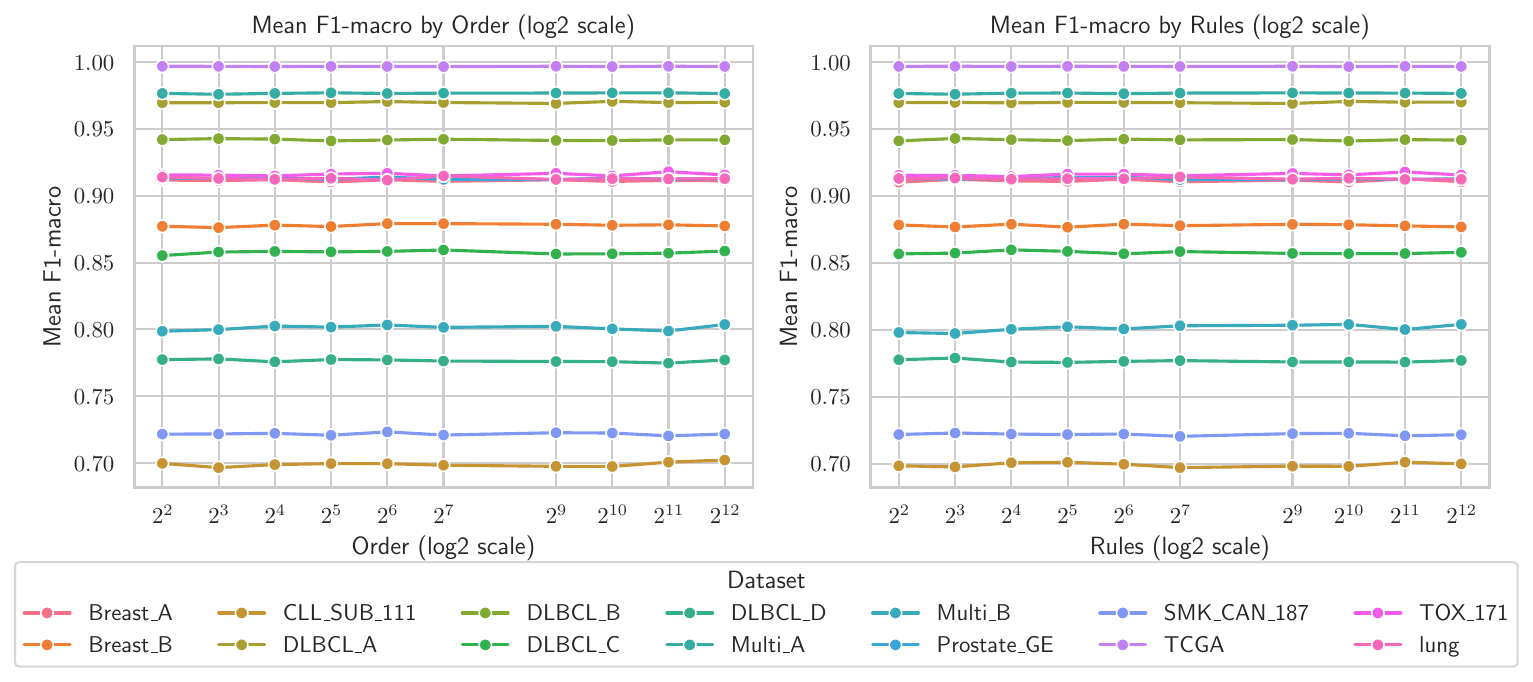}}
    \caption{Interaction of different order and number of rules at different learning rates.}
    \label{fig:heatmap-interact-line}
\end{figure}

Finally, we present the sensitivity analysis, aggregated by the effects of ``Order'' and ``Rules'' across different datasets, as shown in Figure \ref{fig:heatmap-interact-line}. Despite significant differences in the scales of ``Order'' and ``Rules'' (ranging from 4 to 4096), the variations in performance are minor. This suggests that the method exhibits a high degree of stability within the datasets.

\section{Discussion and conclusions}
\label{sec:discussion}
In this section, we discuss the potential impact of the proposed HorNets based on their performance in synthetic and real-world problems.
First, we review the results of the synthetic datasets. This evaluation aimed to empirically test the functional completeness of the proposed HorNets PolyClip capability. The results in Table \ref{tab:res_synth} show that the network can learn any interaction between any number of variables for four neural logic operators: AND, NOT, OR, XNOR, and  XOR. In contrast, the newer variants of neural networks -- TabNET and MLP - as well as the naive rule learning approaches -- the decision trees and random forests - were unable to learn. Since we can learn these combinations of features, we believe that the routing functionality means that we can learn arbitrary combinations of features that interact with each other. This, in turn, shows the potential for using the HorNets applications in many neurosymbolic systems that Garcez et al. \cite{garcez2023neurosymbolic} see as the next generation of AI. On the other hand, \citet{NEURIPS2023_e560202b} emphasizes the problem of linking neurosymbolic learners and considers the production of rules about false relations as linking inferences. We show that the HorNets achieve maximum log-likelihood on both training and test data and thus successfully reconstruct the concept of ground truth -- in the case of logical operations.
\par Using the results on the real biomedical datasets, we establish the effectiveness of the learned rules for subsequent application to real classification problems. The statistical tests confirm that the method can keep up with the strong AutoML baseline (TPOT), while generating interpretable rules that experts can use as a basis for further analyses. This experiment shows that our model can a) train neural networks on small, large tabular datasets, \cite{grinsztajn2022tree} and b) achieve interpretable results, something that related work highlights as necessary \cite{malik2021ten,wong2023explainable,wong2024discovery}
Furthermore, we show that the proposed HorNets are both data- and resource-intensive, making them ideal for applications related to computing on the edge.
We presented a novel approach for learning rules using neural networks. Our method learns from both continuous and discrete signals, as evidenced by rigorous evaluations on both synthetic and real biomedical datasets. The architecture excels at obtaining arbitrary Horn clauses by combining a custom polynomial-like activation function with an attention mechanism that allows it to model complex interactions. This method outperforms advanced AutoML-based learners and is well suited for numerical and categorical data with large table density, while requiring significantly less computational resources than comparable architectures. Our results show that deep neural networks augmented with attention mechanisms and user-defined activation functions are very effective in learning rules.
We plan to improve our framework with uncertainty-based principles to enable the extraction of probabilistic rules. We also plan to extend our approach to explore spaces with mixed inputs. To further improve the interpretability of the extracted rules, we will perform in-depth analyses with domain experts.

\section{Availability and Reproducibility}

The code for the HorNets is freely available via the following GitHub link: \url{https://github.com/bkolosk1/hornets}. To ensure reproducibility and replicability, we provide a Singularity container environment, which integrates the code and provides a benchmarking environment. 

\section{Acknowledgements}

The authors acknowledge the financial support from the Slovenian Research and
Innovation Agency through research the funding schemes P2-0103 and PR-12394.





\bibliography{sn-bibliography}

\appendix

\section*{Appendix: Proof of the Exponentiality of the exact CatInt interactions Using Stirling's Approximation}
\label{app:proof}
Stirling's approximation states that for large $ n $:
\[
n! \approx \sqrt{2 \pi n} \left(\frac{n}{e}\right)^n
\]

Given the complexity expression for the combinatorial part of the CatRouter $\mathcal{O}(\textrm{Comb}(|I|, \textrm{order}))$, let us consider the combination $\textrm{Comb}(|I|, \textrm{order})$ which represents the binomial coefficient:

\[
\textrm{Comb}(|I|, \textrm{order}) = \binom{|I|}{\textrm{order}} = \frac{|I|!}{\textrm{order}! \cdot (|I| - \textrm{order})!}
\]

Using Sterling's approximation for factorials, we have:

\[
|I|! \approx \sqrt{2 \pi |I|} \left(\frac{|I|}{e}\right)^{|I|}
\]

\[
\textrm{order}! \approx \sqrt{2 \pi \, \textrm{order}} \left(\frac{\textrm{order}}{e}\right)^{\textrm{order}}
\]

\[
(|I| - \textrm{order})! \approx \sqrt{2 \pi (|I| - \textrm{order})} \left(\frac{|I| - \textrm{order}}{e}\right)^{|I| - \textrm{order}}
\]

Substituting these into the binomial coefficient formula, we get:

\[
\binom{|I|}{\textrm{order}} \approx \frac{\sqrt{2 \pi |I|} \left(\frac{|I|}{e}\right)^{|I|}}{\sqrt{2 \pi \, \textrm{order}} \left(\frac{\textrm{order}}{e}\right)^{\textrm{order}} \cdot \sqrt{2 \pi (|I| - \textrm{order})} \left(\frac{|I| - \textrm{order}}{e}\right)^{|I| - \textrm{order}}}
\]

\[
\binom{|I|}{\textrm{order}} \approx \frac{\sqrt{2 \pi |I|} \left(\frac{|I|}{e}\right)^{|I|}}{\sqrt{2 \pi \, \textrm{order}} \sqrt{2 \pi (|I| - \textrm{order})} \cdot \left(\frac{\textrm{order}}{e}\right)^{\textrm{order}} \cdot \left(\frac{|I| - \textrm{order}}{e}\right)^{|I| - \textrm{order}}}
\]

\[
\binom{|I|}{\textrm{order}} \approx \frac{\sqrt{2 \pi |I|} }{\sqrt{2 \pi \, \textrm{order}} \sqrt{2 \pi (|I| - \textrm{order})}} \cdot \frac{\left(\frac{|I|}{e}\right)^{|I|}}{  \left(\frac{\textrm{order}}{e}\right)^{\textrm{order}} \cdot \left(\frac{|I| - \textrm{order}}{e}\right)^{|I| - \textrm{order}}}
\]

We can simplify the first term as follows:

\[
\frac{\sqrt{2 \pi |I|}}{\sqrt{2 \pi \, \text{order}} \cdot \sqrt{2 \pi (|I| - \text{order})}} = \frac{\sqrt{2 \pi |I|}}{\sqrt{(2 \pi)^2 \cdot \text{order} \cdot (|I| - \text{order})}} = \frac{\sqrt{2 \pi |I|}}{2 \pi \sqrt{\text{order} \cdot (|I| - \text{order})}}
\]

Next, the exponential term simplifies as follows:

\[
\frac{\left(\frac{|I|}{e}\right)^{|I|}}{\left(\frac{\text{order}}{e}\right)^{\text{order}} \cdot \left(\frac{|I| - \text{order}}{e}\right)^{|I| - \text{order}}} = \frac{|I|^{|I|} \cdot e^{-|I|}}{\text{order}^{\text{order}} \cdot e^{-\text{order}} \cdot (|I| - \text{order})^{|I| - \text{order}} \cdot e^{-(|I| - \text{order})}}
\]

Upon canceling the exponential terms \( e^{-|I|} \), the expression reduces to:

\[
\frac{|I|^{|I|}}{\text{order}^{\text{order}} \cdot (|I| - \text{order})^{|I| - \text{order}}}
\]

The simplified combined expression is:

\[
\binom{|I|}{\textrm{order}} \approx \frac{\sqrt{2 \pi |I|}}{2 \pi \sqrt{\text{order} \cdot (|I| - \text{order})}} \cdot \frac{|I|^{|I|}}{\text{order}^{\text{order}} \cdot (|I| - \text{order})^{|I| - \text{order}}}
\]

Since we are concerned with the asymptotic behavior, we can safely ignore the first term and focus on the latter, thereby demonstrating that the number of combinations in the CatInt grows exponentially.

\[
\mathcal{O}(\textrm{Comb}(|I|, \textrm{order})) \approx \frac{|I|^{|I|}}{\text{order}^{\text{order}} \cdot (|I| - \text{order})^{|I| - \text{order}}}
\]

\end{document}